\documentclass{article}
\usepackage{adjustbox}
\usepackage{booktabs}
\usepackage{url}

\usepackage{graphicx}
\usepackage{amsfonts}
\usepackage{amsmath}
\usepackage{nccmath}
\usepackage{subcaption}
\usepackage{multirow}
\usepackage{multicol}
\usepackage[normalem]{ulem}

\usepackage{amssymb}
\usepackage{amsmath}
\usepackage{graphicx}
\usepackage[colorlinks=true, allcolors=blue]{hyperref}
\usepackage{adjustbox}
\usepackage{authblk}
\usepackage[english]{babel}

\usepackage[letterpaper,top=2cm,bottom=2cm,left=3cm,right=3cm,marginparwidth=1.75cm]{geometry}

\usepackage{amsmath}
\usepackage{graphicx}
\usepackage[colorlinks=true, allcolors=blue]{hyperref}

\title{Kriging and Gaussian Process Interpolation for Georeferenced Data Augmentation}

\author[1,2]{Frédérick Fabre Ferber}
\author[1]{Dominique Gay}
\author[2]{Jean-Christophe Soulie}
\author[1]{Jean Diatta}
\author[3]{Odalric-Ambrym Maillard}

\affil[1]{LIM EA2525, Université de La Réunion}
\affil[2]{UPR Recyclage et risque, CIRAD}
\affil[3]{SCOOL, INRIA}
\date{}

\begin{document}
\maketitle
\begin{abstract}
Data augmentation is a crucial step in the development of robust supervised learning models, especially when dealing with limited datasets. This study explores interpolation techniques for the augmentation of geo-referenced data, with the aim of predicting the presence of Commelina benghalensis L. in sugarcane plots in La Réunion. Given the spatial nature of the data and the high cost of data collection, we evaluated two interpolation approaches: Gaussian processes (GPs) with different kernels and kriging with various variograms. 
The objectives of this work are threefold: (i) to identify which interpolation methods offer the best predictive performance for various regression algorithms, (ii) to analyze the evolution of performance as a function of the number of observations added, and (iii) to assess the spatial consistency of augmented datasets. The results show that GP-based methods, in particular with combined kernels (GP-COMB), significantly improve the performance of regression algorithms while requiring less additional data. Although kriging shows slightly lower performance, it is distinguished by a more homogeneous spatial coverage, a potential advantage in certain contexts.

\end{abstract}

\section{Introduction}

In the field of data-centric artificial intelligence, data augmentation plays a crucial role in the development of training sets. This process consists in enriching the training set by generating new observations without directly collecting additional data. In the literature, two key terms emerge: Data Augmentation (DA) and Data Generation (DG). These concepts cover several practices, such as the augmentation of observations, the creation of new variables, or the treatment of missing values \cite{cui2024tabular}. In this study, we focus on observation augmentation, or data augmentation, in the context of tabular data.
Among existing approaches, rule-based methods extract specific patterns from the data to generate new observations. Variational autoencoders (VAE) combine unsupervised learning and probabilities to generate new data \cite{zhu_s3vae_2020, wan_variational_2017, lotfollahi2020conditional}. Other techniques include generative adversarial networks (GANs) \cite{ouyang2023missdiff, zheng2022diffusion, kotelnikov2023tabddpm}, diffusion models \cite{kotelnikov2023tabddpm, ouyang2023missdiff}, and reinforcement learning \cite{esnaashari2021automation, li2021rule, yu_data_2022}.
In our case, we address a specific problem linked to data augmentation in a game aimed at predicting the presence of the Commelina benghalensis L. species in sugarcane plots on Reunion Island. Collecting these data is time-consuming and costly, and requires specific expertise to guarantee their quality. These constraints considerably limit the volume of data available, in contrast to contexts where the above-mentioned approaches are commonly used. In addition, our data have a spatial dimension, as each plot is geographically located on the island.
To meet these challenges, we are exploring interpolation methods adapted to geo-referenced data, notably kriging and Gaussian processes (GPs). Kriging is a geostatistical method that estimates the values of unsampled points using a variogram to model spatial dependence. This approach is based on the idea that spatially close points have correlated values. Conversely, GPs define a probabilistic distribution over the possible functions, enabling flexible and generalized interpolation.
The aim of this study is threefold: (i) to assess whether these approaches can significantly improve predictive performance by increasing the number of observations available, (ii) to analyze the evolution of performance as a function of the number of observations added, and (iii) to examine whether these methods preserve the spatial consistency of the initial data.
In the remainder of this article, we will detail the concepts of kriging and GPs in section 2. Section 3 will present the methodology and experimental protocol implemented to answer these questions. The results will be analyzed and discussed in section 4, before concluding in section 5.

\section{Kriging and GPs for interpolation} 

Interpolation refers to a model's ability to predict values for data points located between existing observations in the training set. Unlike extrapolation, which concerns predictions outside the range of known data, interpolation focuses on predictions within this range. It plays a crucial role in many algorithms, enabling continuous values of a target variable to be estimated as a function of input variables. For, a training dataset $D = \{ (x_i,y_i) | i= 1,2,\dots, n \}$ and $y_i = f(x_i)$ a scalar for an unknown function $f$ we wish to interpolate the value $y^\star$ of a test point $x^\star$ through $f$. Kriging and GPs are two variants of the same interpolation method, with kriging being the term traditionally used in geo-statistics and GP for machine-learning. Although they refer to the same family of algorithms, there are a few differences, notably in the function used to express the co-variance between the data (one the kernels and the other the variograms) and also in the estimation method used to predict $y^\star$.

\paragraph{Kriging} 

Kriging, named after the South African geostatistician Danie Krige, is a geostatistical interpolation method widely used to estimate the values of unsampled points from existing spatial data. It is based on the theory of regionalized variables and uses variograms to model the spatial dependency structure between data points. 
For a variable of interest at a spatial position $x^\star$, the aim of kriging is to predict $f(x^\star)$ using a linear combination of the observed values $y_i$ for points $x_1 , \dots x_N$ :

$$ f(x^\star) - \mu(x^\star) = \sum_{i=1}^N \lambda_i [ y_i - \mu(x_i)] $$

where \( \lambda_i \) are the weights assigned to each observation $x_i$, $\mu(x_i)$ the mean of the observation points so as to minimize the mean-squared prediction error. Several types of kriging exist, depending on what assumptions can be made about the mean $\mu(x^\star)$: simple kriging assumes that the mean is known, ordinary kriging assumes that the mean is unknown and is therefore estimated locally on the basis of the observation points, and co-kriging uses several correlated variables in addition to spatial variables to improve predictions.
To calculate the weights $\lambda_i $, we use a variogram (or semi-variogram) which describes the variation of $f(x)$ as a function of the distance between two points \\cite{oliver2015basic}.
The variogram is crucial in kriging, as it provides information on the weighting of observations for estimating values at unsampled points: 

$$\begin{bmatrix}
\lambda & \dotsc  & \Sigma_{N1}\\
\vdots  & \Sigma_{ij} & \vdots \\
\Sigma_{1N} & \dotsc  & \Sigma_{NN}
\end{bmatrix} \ \begin{bmatrix}
\lambda _{1}\\
\vdots \\
\lambda _{N}
\end{bmatrix} \ =\ \begin{bmatrix}
\Sigma_{x^\star 1}\\
\vdots \\
\Sigma_{x^\star N}
\end{bmatrix}$$

where $\Sigma_{i,j}$ is the covariance between the observed points $x_i,x_j \in X$ with $i,j \in \{1,\dots,N \}$ and $\Sigma_{x^\star i}$ the covariance between the point $x^\star$ to be interpolated and an observation point $x_i$.

\paragraph{Gaussian Process Regression}
Gaussian Processes (GP) \cite{williams2006gaussian} are non-parametric models defined as a collection of random variables $X$ such that for any finite set of points \( x_1, x_2, \ldots, x_n \), the vector \( (f(x_1), f(x_2), \ldots, f(x_n)) \) follows a multivariate normal distribution. A GP is fully specified by its mean function \( m(X) = \mathbb{E}[f(X)] \) and its covariance function \( k(x, x') = \text{Cov}(f(x), f(x')) \). It estimates the value $f(x^\star)$ such that : 

$$ f(x^\star) \sim \mathcal{GP}(m(x) , k(x,x'))$$

As observations are made, this a priori can be sequentially updated to become an a posteriori distribution of the true function representing the target variable of interest. For regression, the GP will act as an interpolator, calculating the joint distribution between the training data and the point(s) to be predicted: 

 $$ \begin{bmatrix}
\mathbf{f}\\
\mathbf{f_\star}
\end{bmatrix} \ \sim \ \mathcal{N} \ \left(\begin{bmatrix}
\mu\\
\mu_\star
\end{bmatrix} \ ,\ \begin{bmatrix}
K & K\star\\
K\star^T & K \star \star 
\end{bmatrix} \ \right)$$ 
with $\mathbf{f}$ containing the set of target values $\{y_1,\dots, y_n\}$ of the training set, $\mathbf{f \star}$ the predicted values on the test set, $\mu$ the mean on the training set, $\mu \star$ the predicted mean on the test set, $K$ the covariance between training points, $K\star$ the covariance between training points and test points and $K \star \star$ the covariance between test points.

\section{Proposed Method}

Interpolation of geo-referenced data can be seen as data augmentation. Indeed, interpolation algorithms well adapted to a spatial framework seem a particularly suitable solution in this case. These interpolation methods can then be used to increase the number of data items when the latter are limited in number. 
We chose to use the kriging interpolation and Gaussian process regression presented in the previous section to increase the number of data available in an agricultural database designed to predict the cover of the weed species \textit{Commelina benghalensis L.} (COMBE) on sugarcane plots in La Réunion. Three questions may then be raised: (i) which interpolation method for increasing the number of data seems the most effective in terms of predictive performance, (ii) at what threshold of added points does predictive performance seem to converge and (iii) how does the distribution of the species (in terms of cover density) evolve for an interpolation method compared to the distribution on the dataset without increase? 

\paragraph{Dataset} 

We use a dataset comprising geo-referenced real floristic survey data from La Réunion, concerning the presence of the weed \textit{Commelina benghalensis L.} on sugarcane plots \cite{fabre-ferber_dataset_2021,laine2024impact} . This dataset faithfully reflects the challenges common in realistic and complex data contexts, characterized by limited observations. The dataset consists of 745 observations and includes 8 variables, both continuous and categorical summarized in the table \ref{tab:data}. Variables such as plot location were removed as they were not used for interpolation. 
For the variable of interest (species cover) is a continuous variable ranging from 0 to 100, with 0 representing zero cover on a plot and 100 representing total species cover \textit{Commelina benghalensis L.}.

\begin{table}[htbp!]
\centering
\begin{tabular}{c c c}
\hline 
Name & Type & Range \\
\hline 
Longitude & Continuous & $[55.23-55.83]$\\
Latitude & Continuous & $[-21.39-21.88]$ \\
Altitude & Continuous & $[0-950]$\\
Average Temperature & Continuous & $[18-28]$ \\
Precipitation & Continuous & $[0-1400]$ \\
Month & Categorical & $[1-12]$ \\
Year & Categorical & $[2002-2024]$\\
Luminance & Continuous & $[900-1890]$ \\
\hline
\end{tabular}
\caption{Description of the different variables present in the dataset for the recovery of the species \textit{Commelina benghalensis L.}}
\label{tab:data}

\end{table}

\paragraph{Interpolation method}

In a classic case of interpolation, where we wish to estimate the value of a geo-referenced point $x^\star$ from observed points, it is defined by its coordinates in a given projection system (in our case, latitude and longitude). These coordinates can be obtained directly from knowledge of the projection system. However, the problem becomes more complex when we wish to integrate other auxiliary variables, such as altitude or rainfall, which cannot be directly deduced from the projection system. A simple approach would be to assign a mean value or a distribution calculated from the observed points. However, these assumptions can introduce significant deviations from reality, compromising the accuracy of the interpolation.

In our approach, when we create a point $x^\star$ to interpolate to increase the number of points in the learning base, we retrieve the various auxiliary variables using the Meteor\footnote{https://smartis.re/METEOR} service, which requires a date and location. To ensure a balanced temporal distribution, we select data covering all periods of the year, with no over-representation of certain months. In addition, we restrict the points to be interpolated to areas where the probability of finding sugarcane plots is high, thus avoiding irrelevant regions. Regarding interpolation methods, we use several kernels for Gaussian processes: a linear kernel (equation \ref{eq:lin}), an RBF kernel (equation \ref{eq:rbf}) and a quadratic kernel (equation \ref{eq:poly}). We also apply a method for finding combinations of kernels described in \cite{duvenaud2014kernel}, where kernels are constructed using various operators:

$$ \mathcal{S} \rightarrow \mathcal{S+B} $$ 
$$\mathcal{S} \rightarrow \mathcal{S\times B} $$
$$\mathcal{B} \rightarrow \mathcal{B'} $$ with $\mathcal{S}$ representing any kernel subexpression and $\mathcal{B}$ a basic kernel. The basic kernels are those described above and the criterion used to select an optimal kernel is the Bayesian information criterion (BIC) \cite{schwarz1978estimating}. Co-kriging, a variant of classical kriging, is used to incorporate other explanatory variables into the analysis. Several variogram models are employed, including linear variogram (equation \ref{eq:lin_kr}), exponential (equation \ref{eq:exp}), Gaussian (equation \ref{eq:gau}) and spherical (equation \ref{eq:sphe}).

\begin{align}
    k(x, x') &= x \cdot x' \label{eq:lin} \\
    k(x, x') & = \exp\left(-\frac{\|x - x'\|^2}{2\sigma^2}\right) 
 \label{eq:rbf} \\
    k(x, x') &= (x \cdot x' + c)^2  \label{eq:poly}
\end{align}

\begin{align}
    \gamma(h) &= C_0 + bh \label{eq:lin_kr} \\
   \gamma(h) &= C_0 + C \left(1 - \exp\left(-\frac{h}{a}\right)\right) 
 \label{eq:exp} \\
    \gamma(h) &= 
\begin{cases}
    C_0 + C \left( \frac{3h}{2a} - \frac{h^3}{2a^3} \right), & \text{si } 0 \leq h \leq a \\
    C_0 + C, & \text{si } h > a
\end{cases} \label{eq:sphe} \\ 
\gamma(h) &= C_0 + C \left(1 - \exp\left(-\frac{h^2}{a^2}\right)\right) \label{eq:gau}
\end{align} 

with $C_0$ the nugget effect, $C$ the bearing, $h$ the distance and $a$ a scaling parameter.

\paragraph{Protocol}

The protocol implemented in this study aims to answer the questions posed at the beginning of this section. We evaluate the predictive performance of several regression algorithms, measured in terms of mean square error (MSE). Algorithms tested include Linear Regression (LR), Ridge Regression (RR), Support Vector Machine Regressor (SVR), Random Forests (RF), Gradient Boosting Regressor (GB), $k$-nearest neighbors (KNN) and Neural Networks (MLP). These performances are compared between the initial dataset and the dataset augmented by different interpolation methods (tab \ref{tab:my_label}). Three experiments are carried out: (i) Performance evaluation on a test set representing 30\% of the data, comparing the different regression algorithms applied to the original dataset and the one augmented with 200 additional points. (ii) Performance analysis of the best-performing algorithm from the first experiment, as a function of an increasing number of points added $\{0,50,\dots,300\}$, where 0 corresponds to the original dataset. (iii) Comparison of density maps in terms of overlap rate, between those obtained with the initial dataset and those generated by the different interpolation methods where we add 300 new points. 
The choice of kernel parameters for the GPs is made by gradient descent of the log likelihood with the \texttt{GPy}\footnote{https://gpy.readthedocs.io/en/deploy/} library. As no automatic methods are available for variogram parameters, these are chosen according to a set of parameters by minimizing the MSE using the \texttt{PyKrige} library. \footnote{https://geostat-framework.readthedocs.io/projects/pykrige/en/stable/} library. The hyper-parameters of the regression algorithms are left by default and are taken from the \texttt{scikit-learn} library. \cite{scikit-learn}.

\begin{table}[htbp!]
\centering
\begin{tabular}{c c}
\hline 
Method & Acronym \\
\hline
Gaussian process with linear kernel & GP-LIN \\
Gaussian process with polynomial kernel & GP-POLY \\
Gaussian process with RBF & GP-RBF kernel \\
Gaussian process with combination of kernels & GP-COMB \\
Kriging with linear variogram & COKR-LIN \\
Kriging with exponential variogram & COKR-EXP \\
Kriging with Gaussian variogram & COKR-GAU \\
Kriging with spherical variogram & COKR-SPHE \\
\hline
\end{tabular}
\caption{Interpolation methods used}
\label{tab:my_label}
\end{table}

\section{Results and interpretation} 

\begin{figure}[th!]
    \centering
    \includegraphics[width=0.9\linewidth]{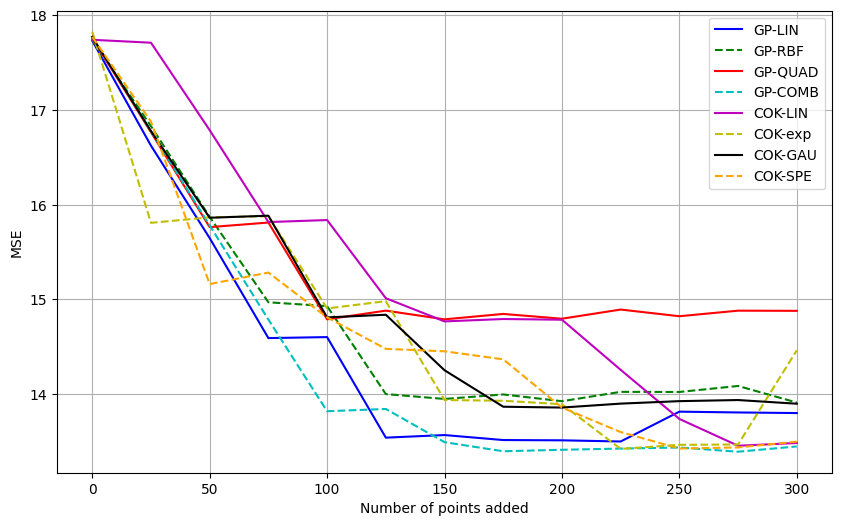}
    \caption{MSE performance of the MLP algorithm as a function of the number of points added by different interpolation techniques. 0 points added corresponds to the original dataset. }
    \label{fig:enter-label}
\end{figure}

\begin{figure}[ht!]
    \centering
    \includegraphics[width=0.99\linewidth]{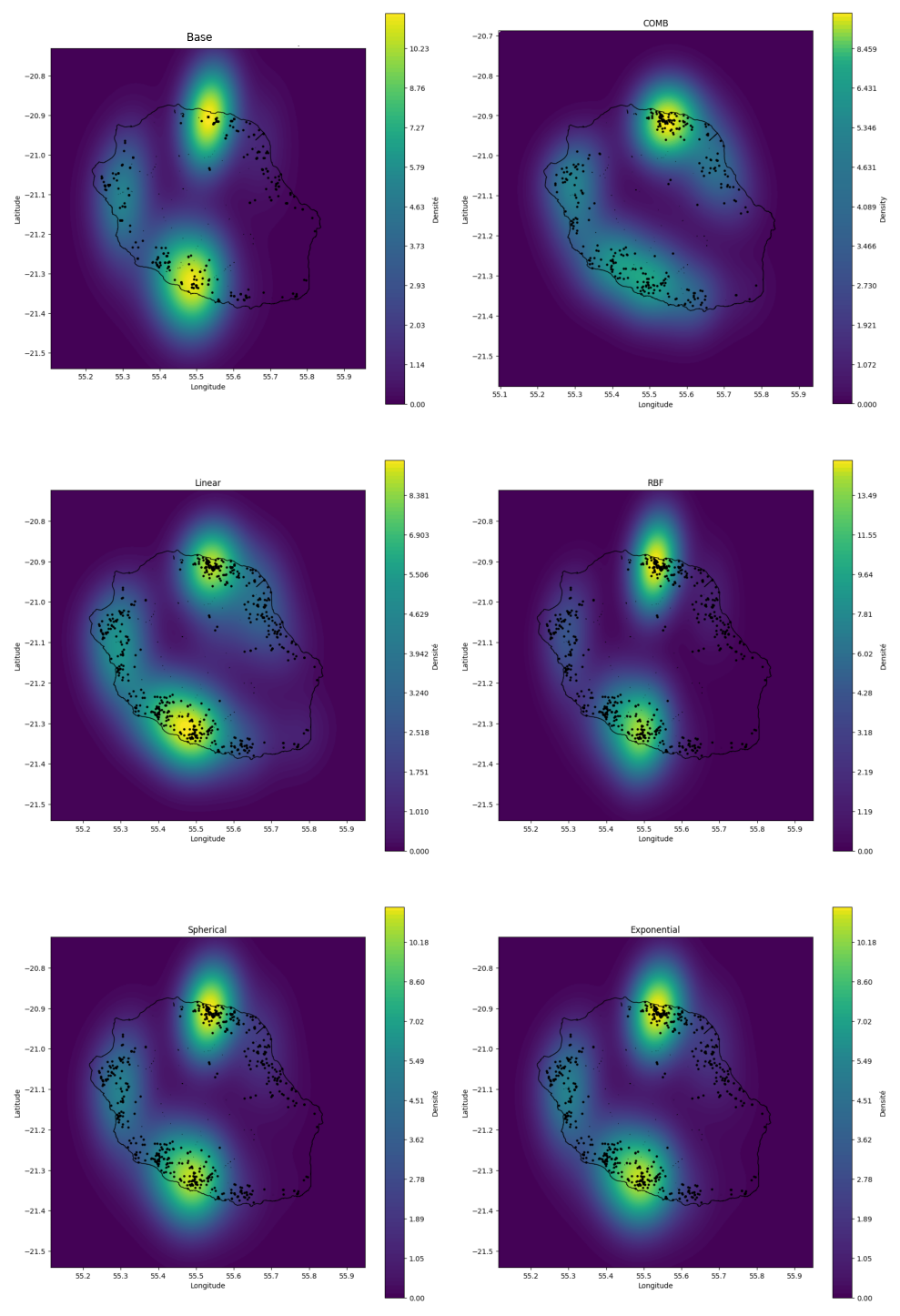}
    \caption{Density map for the species \textit{Commelina benghalensis L.} (COMBE) for the base dataset (Base) and the datasets augmented by 5 interpolation methods: Comb (Combination of kernels), Linear (Linear kernel), RBF (RBF kernel), Exponential Co-Kriging (Exponential) and Spherical Co-Kriging (Spherical).} 
    \label{fig:density}
\end{figure}

This section presents and analyzes the results of various data augmentations via the interpolation techniques described above. 
The results in Table 2 present the root mean square errors (RMSE) for various regression algorithms applied to data sets augmented via different interpolation techniques. Figure \ref{fig:enter-label} shows the evolution of performance for a particular algorithm several number of points added by the different interpolations. Finally, figure \ref{fig:density} shows density maps made with the base dataset and the datasets augmented by the different interpolation techniques.

\begin{table}[htbp]

    \label{tab:results}
    \begin{adjustbox}{width=1.2\textwidth,center}
    \begin{tabular}{lccccccccc}
        \hline
        
        \textbf{Model} & \textbf{Base} & \textbf{GP-RBF} & \textbf{GP-LIN} & \textbf{GP-QUAD} & \textbf{GP-COMB} & \textbf{CoK-LIN} & \textbf{CoK-EXP} & \textbf{CoK-GAU} & \textbf{CoK-SPHE} \\
        \hline
        \textbf{LR} & 18.80 & 13.68 & 13.71 & \textbf{13.67} & 14.57 & 14.58 & 14.78 & 14.66 & 14.56 \\
        \textbf{RR} & 14.84 & 13.68 & 13.71 & \textbf{13.67} & 13.57 & 14.57 & 14.58 & 14.66 & 14.56 \\
        \textbf{SVR} & 16.80 & 14.09 & \textbf{13.74} & 13.93 & 14.98 & 14.87 & 15.12 & 15.11 & 15.11 \\
        \textbf{RF} & 23.83 & 14.05 & 14.05 & 14.01 & \textbf{13.35} & 13.35 & 13.62 & 13.53 & 13.55 \\
        \textbf{BG} & 36.45 & 13.65 & 13.55 & 13.57 & \textbf{13.27} & 13.34 & 13.28 & 13.28 & 13.34 \\
        \textbf{KNN} & 16.30 & 14.44 & 14.55 & 14.44 & \textbf{13.17} & 13.18 & 13.21 & 13.17 & 13.17 \\
        \textbf{MLP} & 17.17 & 13.47 & 13.55 & 13.54 & \textbf{13.41} & 13.49 & 13.47 & 13.44 & \textbf{13.38} \\
        \hline
        
    \end{tabular}
    \end{adjustbox}
        
        \caption{Results of regression models with different interpolation methods for 200 points generated in terms of MSE. Each column represents the interpolation model used, Base represents the dataset without augmentation, and each row represents a regression algorithm. The best performances are marked in bold. }
    \end{table}

\paragraph{Evaluation} The results show that data augmentation via Gaussian Processes and Kriging systematically improves the performance of regression models compared with Base. Among the techniques tested, GP-COMB emerged as the best-performing method for RF, GB and K-NN, showing that it appears to be effective in generating new data that serves performance well. GP-QUAD stands out for LR and RR, effectively modeling both linear and non-linear relationships, making it particularly well-suited to these models.
For kriging interpolation, the Co-K-SPHE and Co-K-EXP variograms produce good results, particularly for MLP and K-NN. However, they show limited compatibility with certain models such as SVR, where performance remains inferior to that of Gaussian processes. The results also reveal that using the dataset without augmentation leads to a noticeable drop in performance, underlining the value of augmentation approaches, particularly in contexts where initial data is limited.
In summary, GP-COMB stands out as the most efficient and versatile method, offering significant improvements for many models but particularly for MLP, which on average out of all interpolation methods has the best performance. However, kriging with the variogram set remains an interesting alternative, especially for models like MLP, and may be preferred depending on context and data characteristics. For the remainder of these experiments, we have chosen to retain the MLP algorithm.

\paragraph{Performance as a function of the number of points added}

Analysis of the evolution of MSE as a function of the number of points added revealed distinct behaviors according to the interpolation methods. GP-COMB and GP-LIN stood out for their rapid convergence towards minimum error, reaching a plateau at around 200 points added. This shows that these methods converge faster, so there's less need to add more points. In contrast, other methods, while effective, generally required a greater number of points to reach a plateau. COK-LIN, for example, showed performance comparable to the best GPs, but with slower convergence. The Gaussian process with quadratic kernel (GP-QUAD) stood out negatively, its error stagnating at a significantly higher level than the other methods. 
Overall, we can see that all the methods increase prediction performance, and that they all stagnate towards the same mean square error, except for GP-RBF, GP-QUAD and COK-GAU, although they are all very close. In conclusion, the addition of extra points generally improves the accuracy of the MLP model. GPs with linear kernels and the combination of kernels, as well as co-kriging with linear and spherical variograms, proved particularly effective in reducing the error as a function of the number of points, but GP-COMB and GP-LIN seem to be favored, stagnating faster around 150-200 points added.

\paragraph{Density map}

\begin{figure}
\centering
\includegraphics[width=0.5\linewidth]{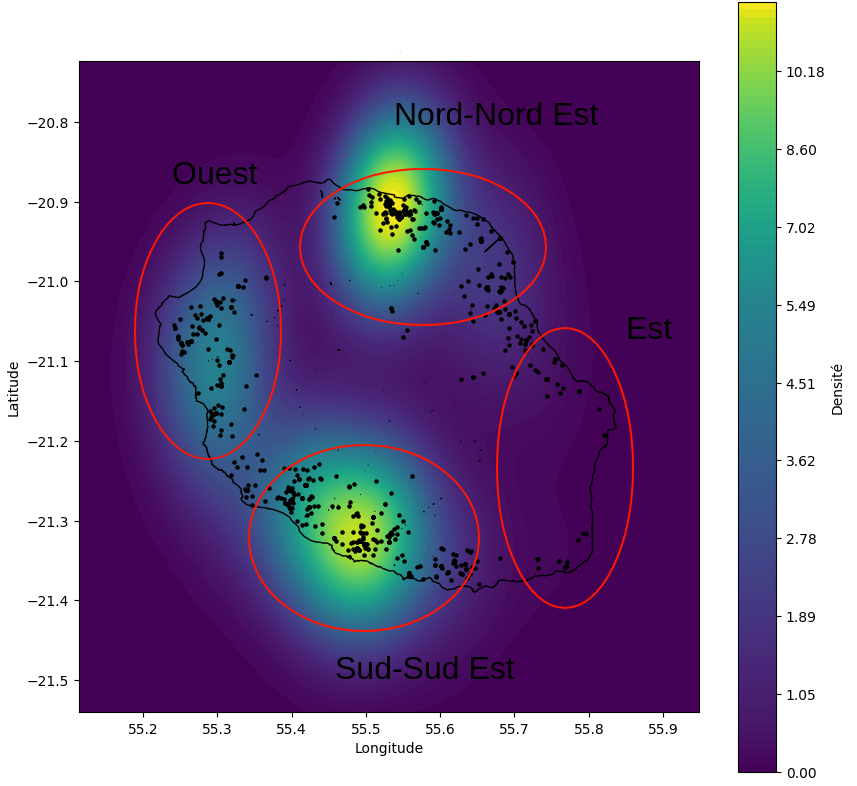}
\caption{Important areas of Reunion Island for the apparition of the species \textit{Commelina benghalensis L.}}
\label{fig:example}
\end{figure}

In this analysis, we compare species cover density maps generated from the original dataset and different interpolation methods, including GP-LIN, GP-RBF, GP-COMB, as well as the COK-SPHE and COK-EXP kriging variants (table \ref{fig:density}). The aim is to assess the extent to which these augmentation methods influence the spatial distribution of the species in different areas of the island: North-North-East, West, South-South-East (figure \ref{fig:example}).

In addition to the various maps, we calculate the average difference in overlap between the points in an area of the original dataset and the areas of the datasets with the added points (table \ref{tab:diff}).

\begin{table}[]
    \centering
    \begin{tabular}{cccccc}
        \hline
       & GP-COMB & GP-LIN & GP-RBF & COK-SPHE & COK-EXP \\
       \hline
         Nord-Nord Est & \textbf{+1.1} &  \textbf{+1.27} & \textbf{+1.9}  & -0.2 & -0.2\\
         Est &  \textbf{+1.56} & \textbf{+1.32} & -0.4 & +0.1 & +0.1  \\
         Ouest & +0.2 & +0.2 & \textbf{-0.8} & -0.14 & -0.14 \\
         Sud-Sud Est & \textbf{-1.2} & -0.24 & \textbf{+0.8}  &  -0.1 & -0.1  \\
         \hline
    \end{tabular}
    \caption{Difference in mean overlap between points in the unaugmented dataset and datasets augmented by the different interpolation techniques. Significant differences are shown in bold.}
    \label{tab:diff}
\end{table}

The GP-COMB and GP-LIN methods are similar in that they accentuate the contrasts in the northern and eastern zones, where the overlap areas are larger than on the base map, which explains the higher mean overlap value. However, GP-COMB tends to represent the species less in the south. GP-RBF, on the other hand, produces more mixed results. It strongly exaggerates the values in the northern and southern zones (negatively), creating very pronounced peaks of cover in the north and slightly less in the south. Although it remains consistent in the western zone like all the other methods, it under-represents the eastern zone of the island with a negative difference. The COK-SPHE and COK-EXP methods offer a more homogeneous view of spatial distribution, with much less marked variations than the other methods.

\paragraph{Summary of results} 
To summarize these results, in terms of predictive performance the GP-COMB and GP-LIN models fare better than kriging models on the validation set. In addition, these kernel methods achieve faster convergence with a reduced number of points. However, analysis of the spatial distribution of overlap reveals some notable differences. Kriging methods maintain a more homogeneous distribution of overlap over the whole island, while kernel methods tend to generalize overlap more between different zones. This difference may be linked to the more flexible nature of the kernels used in GP methods. Nevertheless, further evaluation in an agronomic context is required to determine whether or not this generalization is beneficial.

\section{Conclusion} 
In this study, we explored various interpolation techniques for the augmentation of geo-referenced data, evaluating their effectiveness through several regression algorithms. We also analyzed their impact on the distribution of the species \textit{Commelina benghalensis L.}. Two main approaches were examined: Gaussian processes (GP) with different kernels and kriging using various variograms. The objectives of this research were (i) to identify the methods offering the best performance on validation, (ii) to analyze the evolution of an algorithm's performance as a function of the number of points added, and (iii) to understand the influence of these approaches on the estimated spatial distribution in relation to the initial dataset.
The results indicate that some methods, notably GP-COMB and GP-LIN, stand out by significantly improving the performance of regression algorithms while requiring fewer points to achieve optimal convergence. Kriging methods, although less efficient overall and with slower convergence, offer a more homogeneous spatial coverage of the species. On the other hand, kernel-based approaches tend to produce more generalized models over the whole island, a result that remains to be further investigated to fully assess its implications.
These conclusions open up several perspectives. On the one hand, it would be relevant to apply these techniques to other geo-referenced datasets in order to verify their robustness and applicability. Secondly, adapting these interpolation methods to multi-label scenarios is an interesting avenue. Indeed, the current study focused on data augmentation for a single species (a target variable), but extension to several variables, simultaneously, could considerably broaden the scope of these approaches.

\bibliographystyle{apalike}
\bibliography{main}

\end{document}